\documentclass[letterpaper]{article} % DO NOT CHANGE THIS
\usepackage{aaai2026}  % DO NOT CHANGE THIS
\usepackage{times}  % DO NOT CHANGE THIS
\usepackage{helvet}  % DO NOT CHANGE THIS
\usepackage{courier}  % DO NOT CHANGE THIS
\usepackage[hyphens]{url}  % DO NOT CHANGE THIS
\usepackage{graphicx} % DO NOT CHANGE THIS
\usepackage{natbib}  % DO NOT CHANGE THIS AND DO NOT ADD ANY OPTIONS TO IT
\usepackage{caption} % DO NOT CHANGE THIS AND DO NOT ADD ANY OPTIONS TO IT
\usepackage{algorithm}
\usepackage{algorithmic}

\usepackage{booktabs}
\usepackage{multirow}
\usepackage{array}

\usepackage[most]{tcolorbox}
\tcbuselibrary{breakable}
\usepackage{enumerate}

\usepackage{amsmath}
\usepackage{amssymb}

\usepackage{newfloat}
\usepackage{listings}
\DeclareCaptionStyle{ruled}{labelfont=normalfont,labelsep=colon,strut=off} % DO NOT CHANGE THIS
\floatstyle{ruled}
\newfloat{listing}{tb}{lst}{}
\floatname{listing}{Listing}
\title{SCALE: Selective Resource Allocation for Overcoming Performance Bottlenecks in Mathematical Test-time Scaling}
\author{
    Yang Xiao\textsuperscript{\rm 1} \quad
    Chunpu Xu\textsuperscript{\rm 1} \quad
    Ruifeng Yuan\textsuperscript{\rm 1} \quad
    Jiashuo Wang\textsuperscript{\rm 1} \quad
    Wenjie Li\textsuperscript{\rm 1}\equalcontrib \quad
    Pengfei Liu\textsuperscript{\rm 2,3}\equalcontrib \quad
}
\affiliations{
    \textsuperscript{\rm 1}The Hong Kong Polytechnic University\quad \textsuperscript{\rm 2}Shanghai Jiao Tong University\quad \textsuperscript{\rm 3}SII
    yang-alan.xiao@connect.polyu.hk
}

\usepackage{bibentry}
\makeatletter
\let\old@maketitle\@maketitle
\def\@maketitle{%
  \def\theauthors{\if T\showauthors@on\@author\else Anonymous submission\fi}
  \newcounter{eqfn}\setcounter{eqfn}{0}%
  \newsavebox{\titlearea}
  \sbox{\titlearea}{
    \let\footnote\relax\let\thanks\relax%
    \setcounter{footnote}{0}%
    \def\equalcontrib{%
      \ifnum\value{eqfn}=0%
        \footnote{Corresponding authors.}%
        \setcounter{eqfn}{\value{footnote}}%
      \else%
        \footnotemark[\value{eqfn}]%
      \fi%
    }%
    \vbox{%
      \hsize\textwidth%
      \linewidth\hsize%
      \vskip 0.625in minus 0.125in%
      \centering%
      {\LARGE\bf \@title \par}%
      \vskip 0.1in plus 0.5fil minus 0.05in%
      {\Large{\textbf{\theauthors\ifhmode\\\fi}}}%
      \vskip .2em plus 0.25fil%
      {\normalsize \affiliations_\ifhmode\\\fi}%
      \vskip 1em plus 2fil%
    }%
  }%
  \newlength\actualheight%
  \settoheight{\actualheight}{\usebox{\titlearea}}%
  \ifdim\actualheight>\titlebox%
    \setlength{\titlebox}{\actualheight}%
  \fi%
  \vbox to \titlebox {%
    \let\footnote\thanks\relax%
    \setcounter{footnote}{0}%
    \def\equalcontrib{%
      \ifnum\value{eqfn}=0%
        \footnote{Corresponding authors.}%
        \setcounter{eqfn}{\value{footnote}}%
      \else%
        \footnotemark[\value{eqfn}]%
      \fi%
    }%
    \hsize\textwidth%
    \linewidth\hsize%
    \vskip 0.625in minus 0.125in%
    \centering%
    {\LARGE\bf \@title \par}%
    \vskip 0.1in plus 0.5fil minus 0.05in%
    {\Large{\textbf{\theauthors\ifhmode\\\fi}}}%
    \vskip .2em plus 0.25fil%
    {\normalsize \affiliations_\ifhmode\\\fi}%
    \vskip 1em plus 2fil%
  }%
}%
\makeatother

\begin{document}

\maketitle

\begin{abstract}
Test-time compute scaling has emerged as a powerful paradigm for enhancing mathematical reasoning in large language models (LLMs) by allocating additional computational resources during inference. However, current methods employ uniform resource distribution across all reasoning sub-problems, creating fundamental bottlenecks where challenging sub-problems receive insufficient attention while routine operations consume disproportionate resources. This uniform allocation creates performance bottlenecks where additional computational resources yield diminishing returns. Inspired by dual-process theory, we propose \textbf{SCALE} (Selective Resource Allocation), a framework that selectively allocates computational resources based on sub-problem difficulty. SCALE operates through four stages: (1) problem decomposition into sequential reasoning sub-problems, (2) difficulty assessment of each sub-problem to distinguish between routine operations and computationally challenging sub-problems, (3) selective processing mode assignment between System 1 for simple sub-problems and System 2 for complex ones, and (4) sequential execution with context propagation. By concentrating resources on challenging sub-problems while processing routine operations efficiently, SCALE achieves substantial performance improvements with superior resource utilization. Extensive experiments demonstrate that SCALE significantly outperforms uniform scaling baselines, achieving accuracy improvements of up to 13.75 percentage points (57.50\% to 71.25\% on AIME25) while reducing computational costs by 33-53\%, representing a major advance in test-time scaling that addresses fundamental limitations of current approaches.
\end{abstract}

\begin{links}
    \link{Code}{https://github.com/XiaoYang66/DualThinking}
    \link{Data}{https://huggingface.co/datasets/YangXiao-nlp/DualThinking}
\end{links}

\begin{figure*}[ht]
    \centering
    \includegraphics[width=1\linewidth]{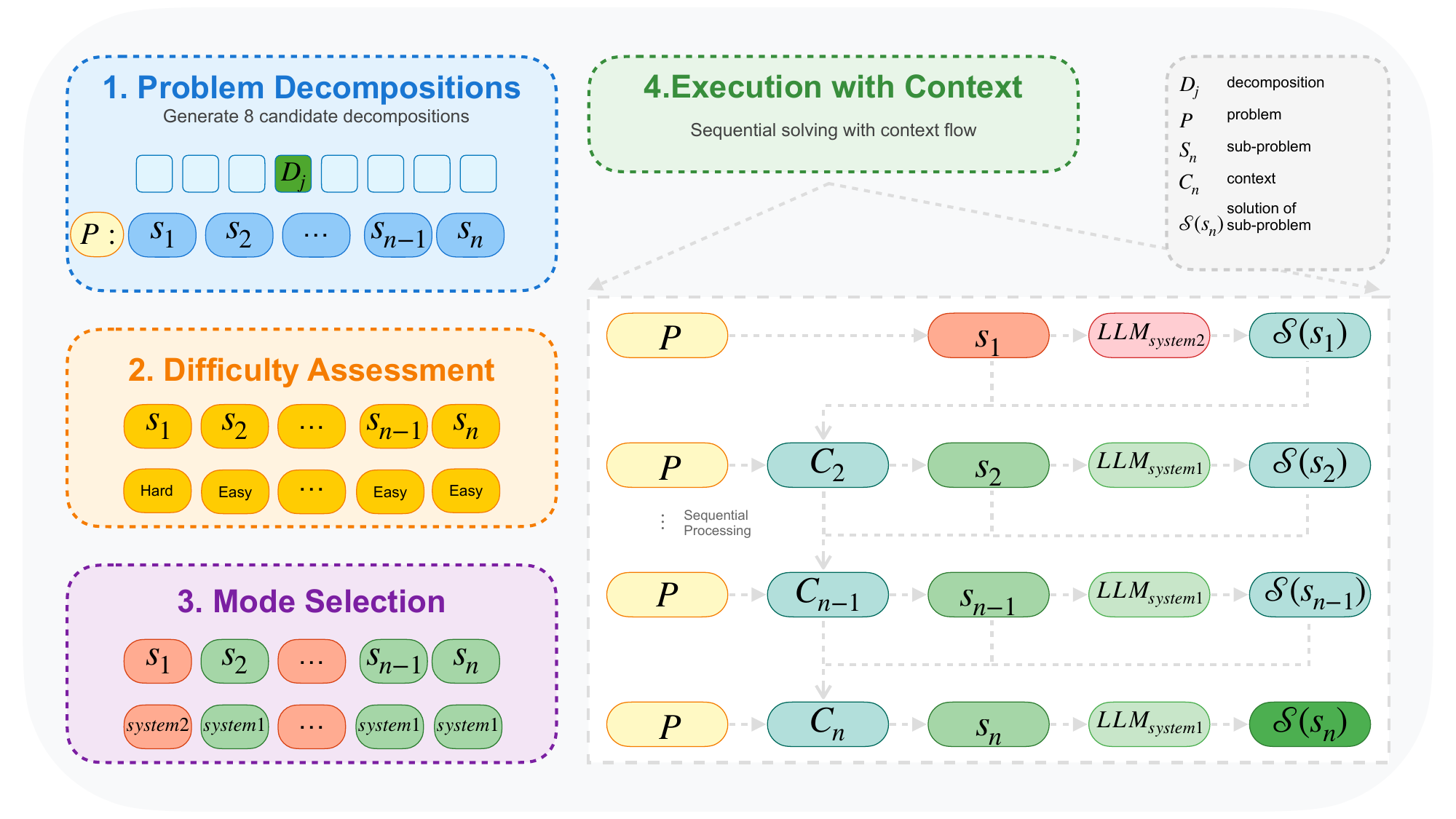}
    \caption{SCALE Framework Overview. SCALE operates through four stages: (1) Problem Decomposition - breaks the mathematical problem into sequential sub-problems; (2) Difficulty Assessment - computes difficulty scores for each sub-problem to distinguish between routine operations and computationally challenging sub-problems; (3) Adaptive Mode Selection - assigns sub-problems to either fast processing (System 1) or deliberate reasoning (System 2) based on difficulty threshold; (4) Sequential Execution - processes sub-problems with full context propagation. This selective resource allocation concentrates computation on challenging sub-problems while efficiently handling routine ones.}
    \label{fig:pipeline}
\end{figure*}

\section{Introduction}

Mathematical reasoning remains a fundamental challenge for large language models, requiring sophisticated multi-step reasoning across diverse cognitive demands. Inference-time compute scaling has emerged as a transformative paradigm that invests additional computational resources during inference rather than relying exclusively on larger model parameters \cite{snell2024scaling,guo2025deepseek}. Recent breakthroughs achieve this scaling by training models to generate extended reasoning traces during inference: supervised fine-tuning approaches like s1 and LIMO train models on long reasoning demonstrations to elicit extended inference-time reasoning, while reinforcement learning methods such as DeepSeek-R1's GRPO algorithm directly incentivize models to develop longer chain-of-thought reasoning patterns through reward-based optimization \cite{muennighoff2025s1,ye2025limo,guo2025deepseek}.

However, existing methods face a fundamental limitation of ``overthinking" in their resource allocation strategies \citep{chen2024not,chiang2024over}. While recent adaptive approaches \cite{shen2025dast,zhang2025adaptthink,yang2025think} have made progress by dynamically adjusting reasoning depth based on global problem difficulty—using techniques such as Token Length Budget metrics and length-aware reward shaping—they still employ uniform resource allocation within individual problems, irrespective of sub-problem complexity.

Consider a mathematical problem containing both routine arithmetic operations and complex algebraic derivations: current methods allocate comparable computational budgets to elementary computations (e.g., $\sqrt{16} = 4$) and intricate sub-problems requiring extensive reasoning tokens. Such uniform allocation creates critical inefficiencies where resources are systematically underutilized on routine operations while challenging solution-determining sub-problems receive insufficient computational attention, leading to performance bottlenecks where additional computational budget yields diminishing returns. The fundamental issue is that even within complex problems, individual reasoning sub-problems can vary significantly in their cognitive demands, ranging from simple computations to advanced mathematical reasoning.

As shown in Figure \ref{fig:pipeline}, we propose SCALE (Selective Resource Allocation), a novel framework that overcomes traditional performance bottlenecks by implementing fine-grained selective resource allocation based on the difficulty of individual reasoning sub-problems. It is motivated by dual-process theory from cognitive science, which distinguishes human cognitive strategies into System 1 (fast, automatic, intuitive) and System 2 (slow, deliberate, effortful) \cite{kahneman2011thinking}. For a reasoning model, our approach operates through four sequential stages that mirror human cognitive processing strategies. (1) \textbf{Problem Decomposition}: the model decomposes the mathematical problem into a sequence of discrete reasoning sub-problems with explicit logical dependencies. (2) \textbf{Difficulty Assessment}: it performs difficulty assessment for each sub-problem, distinguishing between routine operations and computationally challenging sub-problems. (3) \textbf{Adaptive Mode Selection}: leveraging existing models' built-in dynamic thinking mode switching capabilities, it dynamically selects System 1 processing for straightforward sub-problems or System 2 processing for complex sub-problems demanding extensive reasoning resources. (4) \textbf{Sequential Execution}: it solves sub-problems sequentially while propagating contextual information, including previous solved sub-problems and corresponding intermediate results, to maintain coherent reasoning chains throughout the solution process.

Extensive experiments across challenging mathematical reasoning benchmarks demonstrate that SCALE achieves substantial improvements over uniform inference-time scaling baselines. SCALE directly enhances the reasoning model's performance by overcoming the bottlenecks inherent in uniform resource allocation, achieving accuracy improvements of up to 13.75 percentage points (from 57.50\% to 71.25\% on AIME25 with Qwen3-32B) while reducing token usage by 33-53\% compared to baseline methods like InftyThink \cite{yan2025inftythink}. Through supervised fine-tuning on SCALE-generated reasoning traces, non-reasoning models learn to adopt SCALE's inference-time scaling approach, significantly enhancing their reasoning capabilities with accuracy improvements of up to 38.93 percentage points (from 24.58\% to 63.51\% on AIME24 with Llama3.3-70B-Instruct).

Our contributions are threefold: (1) We identify the sub-problem-level resource allocation bottleneck in current inference-time scaling approaches, where uniform allocation prevents effective scaling. (2) We propose SCALE, a cognitively-inspired framework that selectively allocates resources based on sub-problem difficulty, concentrating computation on solution-critical reasoning ones. (3) We demonstrate SCALE's versatility across both enhancing existing reasoning models and enabling non-reasoning models to acquire deep thinking capabilities, achieving substantial performance improvements in mathematical reasoning tasks.

\section{Method}

\subsection{SCALE Framework Overview}

SCALE (Selective Resource Allocation) addresses the fundamental bottleneck of uniform resource allocation in current test-time scaling approaches by implementing selective resource distribution based on sub-problem difficulty. The framework operates through four sequential stages: (1) decomposing mathematical problems into logically ordered sub-problems, (2) evaluating difficulty by distinguishing routine operations from complex reasoning sub-problems, (3) dynamically assigning System 1 processing for simple sub-problems and System 2 processing for challenging ones based on difficulty thresholds, and (4) executing sub-problems sequentially while maintaining complete contextual information to ensure coherent reasoning chains.

\subsection{Mathematical Formulation}

Let $P$ denote a mathematical problem to be solved. SCALE processes $P$ through four sequential stages that work collaboratively to achieve optimal resource allocation. 

\subsubsection{Problem Decomposition}

The decomposition stage transforms the original problem $P$ into a sequence of $n$ logically ordered reasoning sub-problems that build systematically toward the final solution:

\begin{equation}
\mathcal{D}(P) = \{s_1, s_2, \ldots, s_n\}
\label{eq:decomposition}
\end{equation}
where $s_i$ represents a specific sub-problem with clearly defined objectives and logical dependencies on previous sub-problems. The decomposition ensures that each sub-problem either builds upon previous results or can be executed independently while contributing meaningfully to the overall solution pathway.

To ensure robustness, SCALE generates multiple alternative decompositions and selects the optimal one:

\begin{equation}
\mathcal{D}^* = \arg\max_{\mathcal{D}_j \in \{\mathcal{D}_1, \mathcal{D}_2, \ldots, \mathcal{D}_k\}} Q(\mathcal{D}_j, P)
\label{eq:decomposition_selection}
\end{equation}
where $\mathcal{D}_j$ represents a different decomposition, and $Q(\mathcal{D}_j, P)$ evaluates the decomposition quality based on logical correctness, clarity, completeness, and relevance to achieving the final solution. In practice, we prompt the LLM to generate multiple alternative decompositions and prompt them to compare these candidates and select the decomposition that demonstrates the highest quality and effectiveness for the given problem.

\subsubsection{Difficulty Assessment}

For each sub-problem $s_i$ in the selected decomposition, SCALE prompts the model to assess the complexity of the sub-problem:

\begin{equation}
d_i = \mathcal{A}(s_i, C_{i}) \in [0, 1]
\label{eq:difficulty_score}
\end{equation}
where $C_{i}$ represents the accumulated context including the original problem $P$ and all previously solved sub-problems $\{s_1, s_2, \ldots, s_{i-1}\}$ with their solutions. The assessment considers computational complexity, mathematical sophistication required, reasoning depth needed, and solution uncertainty.

\subsubsection{Adaptive Mode Selection}

Modern large language models, including the Qwen3 series~\cite{yang2025qwen3} and Claude family~\cite{claude3}, support switching between different thinking modes. SCALE utilizes this mode-switching capability by dynamically selecting the appropriate processing mode based on sub-problem difficulty.

Based on difficulty assessment, SCALE assigns each sub-problem to one of two LLM processing modes:

\begin{equation}
m_i = \begin{cases}
\text{LLM}_{\text{system1}} & \text{if } d_i \leq \tau \\
\text{LLM}_{\text{system2}} & \text{if } d_i > \tau
\end{cases}
\label{eq:mode_selection}
\end{equation}
where $m_i$ represents the processing mode assigned to sub-problem $s_i$, and $\tau$ represents the difficulty threshold that determines the boundary between routine (easy) and complex (hard) sub-problems. For models supporting dual processing modes (e.g., Qwen3), System 1 mode utilizes the model's efficient processing capabilities for direct computation, while System 2 mode activates deliberate reasoning with chain-of-thought processing and comprehensive analysis. For reasoning-only models (e.g., QwQ~\cite{qwq32b}), we pair them with Qwen3's System 1 counterpart to ensure fair comparison. This threshold-based selection ensures that computational resources are allocated based on the sub-problem's actual complexity rather than assumptions about problem difficulty.

\subsubsection{Sequential Execution with Context Propagation}

SCALE processes sub-problems sequentially while maintaining coherent contextual information throughout the solution process. For each sub-problem $s_i$, the system constructs a comprehensive context that includes the original problem, all previous sub-problems with their corresponding solutions:

\begin{equation}
C_i = \begin{cases}
P & \text{if } i = 1 \\
P \cup \bigcup_{j=2}^{i} \{s_j, \mathcal{S}(s_j)\} & \text{if } i \geq 2
\end{cases}
\label{eq:context_construction}
\end{equation}
where $\mathcal{S}(s_j)$ represents the solution or reasoning output generated for sub-problem $s_j$. This ensures that each sub-problem is solved with full awareness of both its specific requirements and the complete reasoning history.

The solution process adapts to the assigned processing mode:

\begin{equation}
\mathcal{S}(s_i) = m_i(C_i,s_i)
\label{eq:solution_process}
\end{equation}

Finally, SCALE extracts the final answer from the solution of the last sub-problem $\mathcal{S}(s_n)$, ensuring that the complete reasoning chain culminates in a coherent solution to the original problem $P$.

\subsection{SCALE Process Formalization}
SCALE's complete process can be formalized as generating the final answer by integrating solutions from all sub-problems through the four-stage pipeline. This process represents a conditional probability that depends on the complete reasoning chain:

\begin{equation}
\mathbb{P}(\text{Answer} | P) = \prod_{i=1}^{n} P(\mathcal{S}(s_i) | C_{i},s_i)
\label{eq:final_answer}
\end{equation}
where the final answer is conditioned on the original problem, the sequential pairs of sub-problems, and their solutions. The specific prompts utilized for each stage of the SCALE framework are provided in the appendix.

\section{Experimental Setup}

We evaluate SCALE through two complementary experimental settings designed to demonstrate its versatility and effectiveness. In the first setting, we examine SCALE as a prompt-based framework that directly enhances the performance of existing reasoning models at inference time. In the second setting, we investigate SCALE's effectiveness in generating high-quality synthetic reasoning traces for improving non-reasoning models through supervised fine-tuning. This setting demonstrates SCALE's versatility as a data generation tool for model training.

\subsection{Setting 1: Enhancing Reasoning Model Performance}

In this setting, we deploy SCALE as a prompt framework to improve the test-time performance of state-of-the-art reasoning models without requiring any additional training. We evaluate SCALE on four advanced reasoning-capable models: Qwen3-32B~\cite{yang2025qwen3}, QwQ~\cite{qwq32b}, DeepSeek-R1-Distill-Llama-70B, and DeepSeek-R1-Distill-Qwen-32B~\cite{guo2025deepseek}. For brevity, we refer to the DeepSeek-R1 distilled variants as Distill-Llama-70B and Distill-Qwen-32B, respectively, in our result tables.

We compare SCALE against three baseline approaches: (1) The first is zero-shot chain-of-thought~\cite{wei2022chain}, denoted as \textbf{CoT}, which serves as the standard baseline using the prompt ``Please reason step by step, and put your final answer within \textbackslash boxed\{\}.'' (2) The second baseline is \textbf{InftyThink}~\cite{yan2025inftythink}, a recent method that transforms monolithic reasoning into an iterative process with intermediate summarization. (3) The third baseline is \textbf{Majority-Voting}~\cite{DBLP:conf/iclr/0002WSLCNCZ23}, which samples multiple reasoning chains from the base model using CoT prompting and selects the most frequent answer among the generated responses. This approach leverages the diversity of multiple reasoning paths to improve robustness and accuracy through ensemble voting.

\subsection{Setting 2: Enhancing Non-Reasoning Model Performance}

The second setting explores SCALE's effectiveness as a framework for generating high-quality synthetic reasoning traces that can be used to enhance non-reasoning models through supervised fine-tuning.

For our experimental setup, we utilize the problem set from the LIMOPro~\cite{xiao2025limopro} and apply our SCALE framework on QwQ to synthesize reasoning traces for these problems. To ensure data quality and consistency, we filter out instances where the final answers generated by SCALE differ from the original LIMOPro answers, resulting in a curated dataset of 800 high-quality question-response pairs with SCALE-generated reasoning traces. We then perform supervised fine-tuning on four base models: Qwen2.5-14B-Instruct, Qwen2.5-32B-Instruct, Qwen2.5-72B-Instruct~\cite{qwen2025qwen25technicalreport}, and Llama3.3-70B-Instruct \citep{llama3modelcard}, using the same training configurations as LIMOPro, including hyperparameters, optimization schedule, and fine-tuning procedures. We use "-I" to denote "-Instruct" for brevity in the table.

\subsection{Evaluation Benchmarks and Metrics}

Both experimental settings are evaluated on three challenging mathematical reasoning datasets: (1) AIME 2024 (AIME24)~\cite{AIME24}, which contains problems from the 2024 American Invitational Mathematics Examination representing high-school competition-level mathematics; (2) AIME 2025 (AIME25)~\cite{AIME25}, featuring the most recent AIME problems to test the models' ability to generalize to unseen mathematical challenges; and (3) AMC 2023 (AMC23)~\cite{AMC23}, comprising problems from the 2023 American Mathematics Competitions that cover a broader range of mathematical topics.

To comprehensively assess both effectiveness and efficiency, we employ three key metrics. \textbf{Acc} measures pass@1 accuracy as the percentage of problems solved correctly, obtained by sampling each model eight times. Since InftyThink involves multiple rounds of iterative reasoning while our approach generates different sub-problems, we introduce \textbf{Tpi} (tokens per iteration) as a metric to fairly compare resource allocation granularity across different methods. Tpi calculates response tokens per iteration or sub-problems. \textbf{Tok} records total response tokens per problem, directly correlating with computational cost and inference time. 

For inference parameters, all models use a temperature of 0.6 to balance creativity and consistency, with top-p set to 0.95 for nucleus sampling. Other settings follow the default configurations of their respective models.

\section{Enhancing Reasoning Model Performance}
We evaluate SCALE's effectiveness in enhancing reasoning model performance. Table  \ref{tab:setting1_results} presents comprehensive results comparing SCALE against other baselines.

\subsection{Main Results}
\label{section:main results}
Table \ref{tab:setting1_results} presents the comprehensive evaluation of SCALE against baseline approaches across three challenging mathematical reasoning benchmarks. The results demonstrate that SCALE's selective resource allocation strategy effectively overcomes the performance bottlenecks inherent in uniform resource distribution methods.

\textbf{Overcoming Performance Bottlenecks.} Compared to CoT baselines, SCALE achieves substantial improvements across all model-dataset combinations, with the most pronounced gains on AIME24 and AIME25. For Qwen3-32B, SCALE improves accuracy by 9.59 percentage points on AIME24 (from 73.33\% to 82.92\%) and 13.75 percentage points on AIME25 (from 57.50\% to 71.25\%). Similar patterns emerge across other models: QwQ shows improvements of 4.40 and 6.88 percentage points on AIME24 and AIME25, respectively. These results demonstrate that SCALE addresses the fundamental bottleneck where uniform distribution prevents effective scaling despite increasing computational resources.

\begin{table*}[ht]
\centering
\small
\setlength{\tabcolsep}{1mm}
\begin{tabular}{l|ccc|ccc|ccc|c}
\toprule
\multirow{2}{*}{\textbf{Model/Method}} & \multicolumn{3}{c|}{\textbf{AIME24}} & \multicolumn{3}{c|}{\textbf{AIME25}} & \multicolumn{3}{c|}{\textbf{AMC23}} & \multirow{2}{*}{\textbf{Avg. Acc (\%)}} \\
\cmidrule(lr){2-4} \cmidrule(lr){5-7} \cmidrule(lr){8-10}
& \textbf{Acc (\%)} & \textbf{Tpi} & \textbf{Tok} & \textbf{Acc (\%)} & \textbf{Tpi} & \textbf{Tok} & \textbf{Acc (\%)} & \textbf{Tpi} & \textbf{Tok} & \\
\midrule
Qwen3-32B-CoT & 73.33 & 7,409 & 7,409 & 57.50 & 6,839 & 6,839 & 96.88 & 5,789 & 5,789 & 75.90 \\
Qwen3-32B-InftyThink & \textbf{83.75} & 5,900 & 38,451 & 70.00 & 6,538 & 36,640 & 98.12 & 3,723 & 23,117 & 83.96 \\
Qwen3-32B-Majority-Voting & 76.67 & / & 59,279 & 53.33 & / & 54,719 & 97.50 & / & 46,315 & 75.83 \\
Qwen3-32B-SCALE & 82.92 & 3,550 & 25,581 & \textbf{71.25} & 3,709 & 26,643 & \textbf{98.44} & 1,991 & 12,556 & \textbf{84.20} \\
\midrule
QwQ-CoT & 75.00 & 9,957 & 9,957 & 63.33 & 11,087 & 11,087 & 96.88 & 6,161 & 6,161 & 78.40 \\
QwQ-InftyThink & 77.25 & 8,722 & 62,557 & 65.53 & 10,114 & 73,473 & 97.19 & 5,449 & 34,725 & 79.99 \\
QwQ-Majority-Voting & 76.67 & / & 79,663 & 63.33 & / & 88,696 & \textbf{97.50} & / & 49,291 & 79.17 \\
QwQ-SCALE & \textbf{79.40} & 4,157 & 29,812 & \textbf{70.21} & 4,673 & 33,965 & 96.25 & 2,551 & 16,259 & \textbf{81.95} \\
\midrule
Distill-Llama-70B-CoT & 70.00 & 8,159 & 8,159 & 61.37 & 8,218 & 8,218 & 95.31 & 4,825 & 4,825 & 75.56 \\
Distill-Llama-70B-InftyThink & 69.58 & 5,921 & 42,143 & 54.58 & 5,362 & 38,763 & 93.75 & 3,844 & 24,327 & 72.64 \\
Distill-Llama-70B-Majority-Voting & \textbf{76.67} & / & 65,277 & 56.67 & / & 63,830 & 95.00 & / & 38,605 & 76.11 \\
Distill-Llama-70B-SCALE & 71.25 & 4,313 & 30,696 & \textbf{62.08} & 4,913 & 35,517 & \textbf{96.88} & 2,566 & 16,238 & \textbf{76.74} \\
\midrule
Distill-Qwen-32B-CoT & 72.50 & 7,909 & 7,909 & 58.75 & 8,699 & 8,699 & 95.00 & 4,994 & 4,994 & 75.42 \\
Distill-Qwen-32B-InftyThink & 67.50 & 5,244 & 37,062 & 48.75 & 4,915 & 35,187 & 92.81 & 3,711 & 23,332 & 69.69 \\
Distill-Qwen-32B-Majority-Voting & \textbf{80.00} & / & 63,276 & \textbf{60.00} & / & 69,595 & 95.00 & / & 39,953 & \textbf{78.33} \\
Distill-Qwen-32B-SCALE & 73.33 & 4,270 & 30,175 & 59.17 & 4,140 & 29,641 & \textbf{95.63} & 2,672 & 16,804 & 76.04 \\
\bottomrule
\end{tabular}
\caption{Performance comparison of SCALE against baseline methods on mathematical reasoning benchmarks. Acc: accuracy (\%), Tpi: tokens per iteration, Tok: total tokens per problem.}
\label{tab:setting1_results}
\end{table*}

\textbf{Superior Performance-Resource Trade-offs.} When compared to InftyThink, SCALE demonstrates substantial performance improvements while achieving remarkable efficiency gains. SCALE consistently outperforms InftyThink on AIME25 across all models, with notable improvements of 10.42 and 7.50 percentage points for DeepSeek variants, and 4.68 percentage points for QwQ. On AIME24, SCALE shows strong gains with DeepSeek models (+5.83 and +1.67 percentage points) and QwQ (+2.15 percentage points). Against Majority Voting approaches, SCALE achieves competitive accuracy while using 2-3× fewer computational resources. This highlights SCALE's ability to match Majority Voting while avoiding the computational overhead that creates diminishing returns in uniform scaling approaches.

\textbf{Enhanced Resource Utilization Through Selective Allocation.} A key insight from our results is that SCALE achieves superior performance while demonstrating significantly improved resource utilization. Examining tokens per iteration (Tpi), SCALE consistently reduces the average computational cost per reasoning step—achieving 3,550 Tpi versus InftyThink's 5,900 Tpi on Qwen3-32B for AIME24, representing a 40\% reduction. This efficiency translates to total token usage (Tok) that is 33-53\% lower than InftyThink across different model-dataset combinations while maintaining superior accuracy. For instance, on AIME25 with Qwen3-32B, SCALE uses only 26,643 total tokens compared to InftyThink's 36,640 tokens while achieving higher accuracy (71.25\% vs 70.00\%). Notably, the Majority Voting approaches consume substantially more resources (54,719-88,696 total tokens) yet fail to consistently outperform SCALE, highlighting the efficiency of selective allocation over brute-force ensemble methods. This enhanced resource utilization directly validates our core hypothesis: concentrating resources where they can maximize impact yields better returns than uniform allocation strategies that create performance bottlenecks.

\textbf{Model-Agnostic Bottleneck Resolution.} The consistency of SCALE's improvements across diverse model architectures—from Qwen3-32B, QwQ, to Llama variants—demonstrates that SCALE addresses fundamental bottlenecks in test-time scaling rather than exploiting model-specific characteristics. Notably, SCALE achieves an average accuracy improvement of 3.41 percentage points over CoT baselines across all models and datasets, with the framework showing particular effectiveness on the more challenging AIME benchmarks where the bottlenecks of uniform resource allocation are most pronounced.

These results establish SCALE as a significant advancement in test-time scaling that can overcome the performance bottlenecks that limit uniform computational distribution approaches.

\subsection{Inference-Time Scaling Properties of SCALE}

\begin{figure}[t]
    \centering
    \small
    \includegraphics[width=\linewidth]{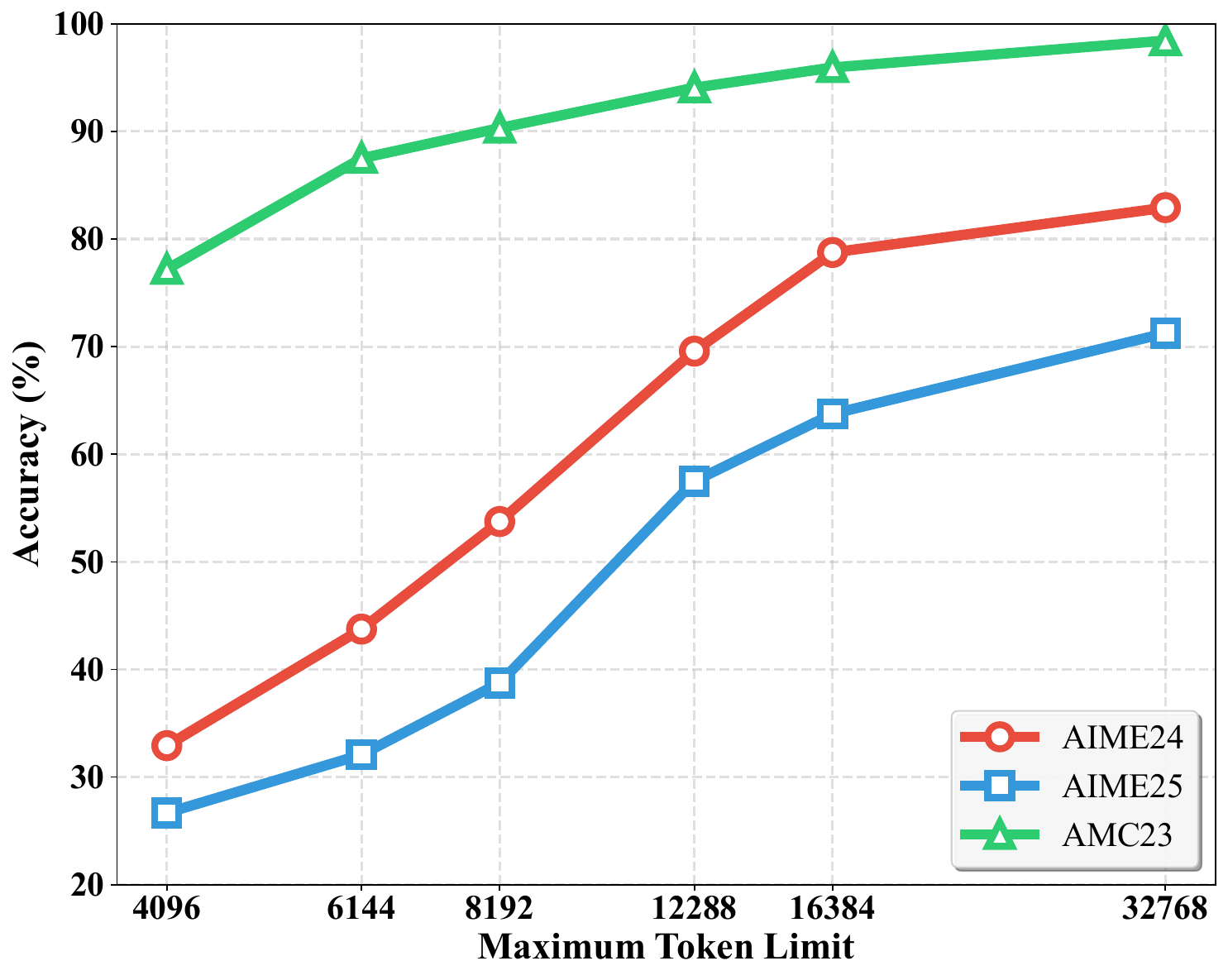}
    \caption{Inference-time scaling of SCALE for Qwen3-32B-SCALE across three benchmarks.}
    \label{fig:scaling}
\end{figure}

To investigate SCALE's adherence to inference-time scaling laws and validate our selective resource allocation hypothesis, we systematically vary the Qwen3-32B's maximum token limit for System 2 processing across different computational budgets ranging from 4,096 to 32,768 tokens. This experimental design allows us to examine how SCALE's performance scales with increased computational resources of System 2 while maintaining the token limit of System 1 processing for routine sub-problems. The result is demonstrated in Figure~\ref{fig:scaling}.

\textbf{Validation of Selective Resource Allocation.} The dramatically different scaling curves between AMC23 and AIME benchmarks provide direct empirical validation of our core analysis about selective resource allocation. AIME24 and AIME25 exhibit steep, almost linear scaling trajectories, while AMC23 shows a more gradual improvement curve that begins to plateau. This divergence directly reflects the underlying distribution of sub-problem difficulties: AIME benchmarks inherently contain a higher proportion of difficult sub-problems, whereas AMC23 problems consist relatively more of routine computational sub-problems. In our experiments, as we increase the maximum token limit, only sub-problems classified as hard receive additional computational resources, while simple sub-problems maintain constant resource allocation. Therefore, when SCALE processes AIME problems, a larger fraction of sub-problems are identified as challenging and benefit from the increased token budget, resulting in steep performance gains. This empirical evidence confirms that concentrating resources on difficult sub-problems---rather than distributing them uniformly---yields superior performance improvements.

\textbf{Adherence to Inference-Time Scaling Laws.} Beyond validating our selective allocation approach, the results demonstrate that SCALE successfully preserves the fundamental scaling law principle---that increased computational resources during inference lead to improved performance. The monotonic improvement across all token budgets (from 4,096 to 32,768 tokens) confirms that SCALE's selective allocation does not disrupt the underlying scaling dynamics but rather amplifies them. By concentrating additional resources precisely where they provide maximum benefit, SCALE achieves more efficient scaling compared to uniform allocation approaches. This is particularly evident in the near-linear scaling observed for AIME benchmarks, where each doubling of the token budget yields consistent accuracy improvements of approximately 10-15 percentage points, demonstrating that selective allocation creates more favorable scaling conditions for problems with heterogeneous cognitive demands.

\begin{table}[t]
\centering
\small
\begin{tabular}{c|ccc|ccc}
\toprule
\multirow{2}{*}{\textbf{Threshold}} & \multicolumn{3}{c|}{\textbf{AIME24}} & \multicolumn{3}{c}{\textbf{AIME25}} \\
\cmidrule(lr){2-4} \cmidrule(lr){5-7}
& \textbf{Acc} & \textbf{Tok} & \textbf{Hard } & \textbf{Acc} & \textbf{Tok} & \textbf{Hard } \\
\midrule
0.2 & 78.75 & 22.5 & 75.61 & 63.75 & 23.2 & 75.78 \\
0.3 & 77.50 & 19.2 & 56.66 & 61.67 & 20.6 & 57.53 \\
0.4 & 73.33 & 19.5 & 55.66 & 51.67 & 20.5 & 57.93 \\
0.5 & 74.58 & 20.3 & 55.68 & 56.25 & 21.0 & 58.53 \\
0.6 & 70.00 & 17.8 & 47.69 & 58.75 & 19.0 & 50.97 \\
0.7 & 50.00 & 12.2 & 17.68 & 39.17 & 11.7 & 19.60 \\
0.8 & 46.25 & 11.4 & 14.08 & 35.83 & 10.6 & 13.95 \\
0.9 & 27.50 & 8.2 & 0 & 25.00 & 6.8 & 0 \\
\bottomrule
\end{tabular}
\caption{Analysis on difficulty threshold values for Qwen3-32B-SCALE across AIME datasets. Acc: accuracy percentage, Tok: total tokens per problem in thousands, Hard: percentage of sub-problems classified as hard.}
\label{tab:ablation_threshold}
\end{table}

\subsection{Difficulty Threshold Analysis}

To understand the impact of the difficulty threshold $\tau$ on SCALE's performance, we conduct a comprehensive analysis using Qwen3-32B across different threshold values ranging from 0.2 to 0.9. The difficulty threshold determines the boundary for classifying sub-problems as easy (processed by System 1) or hard (processed by System 2), making it a critical hyperparameter for optimal resource allocation. Table~\ref{tab:ablation_threshold} presents the results.

The results show a clear trend: lower thresholds consistently achieve higher accuracy across both datasets. Optimal performance is achieved at $\tau = 0.2$, reaching 78.75\% on AIME24 and 63.75\% on AIME25, with accuracy systematically decreasing as the threshold increases. At $\tau = 0.2$, approximately 75\% of sub-problems are classified as hard and processed through System 2, indicating that allocating more sub-problems to System 2 substantially benefits mathematical problem-solving, even at the cost of increased computation.

Notably, even with higher thresholds that reduce accuracy, SCALE maintains substantial advantages over CoT baselines. For instance, on AIME24, SCALE with $\tau = 0.5$ achieves 74.58\% accuracy—still outperforming CoT's 73.33\%. On AIME25, SCALE at $\tau = 0.6$ (58.75\%) exceeds CoT performance (57.5\%). This demonstrates SCALE's robustness: even suboptimal threshold settings preserve competitive performance while requiring significantly fewer sub-problems (47.69\% and 50.97\% respectively) to undergo System 2 processing.

The inverse relationship between threshold and token usage presents practical deployment opportunities. As $\tau$ increases from 0.2 to 0.9, total tokens decrease dramatically—by 63\% on AIME24 (from 22,454 to 8,229 tokens) and 71\% on AIME25 (from 23,212 to 6,811 tokens). Correspondingly, the percentage of hard sub-problems decreases from over 75\% at low thresholds to 0\% at $\tau = 0.9$. This flexibility allows practitioners to adjust $\tau$ based on computational budgets—using lower thresholds when accuracy is paramount, or higher thresholds when operating under strict resource constraints while still maintaining advantages over baseline methods.

\section{Enhancing Non-Reasoning Model Performance}

In addition to enhancing reasoning-capable models, we investigate SCALE's effectiveness in generating high-quality synthetic reasoning traces for improving non-reasoning models through supervised fine-tuning. This setting demonstrates SCALE's versatility as a data generation tool for model training.

Table~\ref{tab:setting2_results} presents the comprehensive comparison between base models and their SCALE-enhanced counterparts across four different models. The results demonstrate substantial and consistent improvements across all model sizes and datasets.

\begin{table}[t]
\centering
\small
\setlength{\tabcolsep}{1mm}
\begin{tabular}{l|cc|cc|cc}
\toprule
\multirow{2}{*}{\textbf{Model}} & \multicolumn{2}{c|}{\textbf{AIME24}} & \multicolumn{2}{c|}{\textbf{AIME25}} & \multicolumn{2}{c}{\textbf{AMC23}} \\
\cmidrule(lr){2-3} \cmidrule(lr){4-5} \cmidrule(lr){6-7}
& \textbf{Acc} & \textbf{Tok} & \textbf{Acc} & \textbf{Tok} & \textbf{Acc} & \textbf{Tok} \\
\midrule
Qwen2.5-14B-I & 13.75 & 1.0 & 13.33 & 0.9 & 57.19 & 0.8 \\
\quad w/ SCALE & \textbf{27.50} & 24.4 & \textbf{20.00} & 25.0 & \textbf{61.88} & 14.5 \\
\midrule
Qwen2.5-32B-I & 17.92 & 0.9 & 16.67 & 0.9 & 68.75 & 0.7 \\
\quad w/ SCALE & \textbf{53.33} & 22.0 & \textbf{39.17} & 22.2 & \textbf{86.23} & 10.7 \\
\midrule
Qwen2.5-72B-I & 17.92 & 1.2 & 12.92 & 1.0 & 66.87 & 0.9 \\
\quad w/ SCALE & \textbf{54.17} & 21.1 & \textbf{47.50} & 21.4 & \textbf{88.12} & 11.7 \\
\midrule
Llama3.3-70B-I & 24.58 & 1.2 & 5.42 & 1.0 & 57.81 & 1.0 \\
\quad w/ SCALE & \textbf{63.51} & 15.7 & \textbf{44.83} & 16.5 & \textbf{83.23} & 10.7 \\
\bottomrule
\end{tabular}
\caption{Performance comparison of SCALE-enhanced models against base models. Acc: accuracy (\%), Tok: total tokens per problem in thousands.}
\label{tab:setting2_results}
\end{table}

\textbf{Significant Performance Gains Across All Benchmarks.} All fine-tuned models show remarkable improvements over their base versions, with gains ranging from 4.69 to 39.41 percentage points across different benchmarks. The most notable improvements occur on challenging AIME problems, where Llama3.3-70B-Instruct achieves dramatic gains of 38.93 percentage points on AIME24 (from 24.58\% to 63.51\%) and Qwen2.5-32B-Instruct improves by 35.41 percentage points on the same benchmark.

\textbf{Cross-Architecture Generalization and Scaling Effects.} SCALE demonstrates strong generalization across different model architectures and sizes. Within the Qwen2.5 family, clear scaling trends emerge: the 32B and 72B models achieve substantially larger gains than the 14B model, particularly on challenging AIME problems. However, the performance gap between 32B and 72B models is relatively small (53.33\% vs 54.17\% on AIME24), suggesting diminishing returns beyond a certain capacity threshold. Importantly, Llama3.3-70B achieves competitive performance (63.51\% on AIME24), demonstrating that SCALE's effectiveness transcends specific architectural choices.

\textbf{Enhanced Reasoning Pattern Transfer.} The consistent improvements across different problem difficulties indicate that SCALE-generated reasoning traces effectively transfer sophisticated reasoning patterns to non-reasoning models. This is particularly evident on challenging AIME problems, where base models initially struggled with accuracy rates below 25\%. After fine-tuning with SCALE-generated data, larger models reach 44-63\% accuracy on AIME benchmarks, representing performance gains that would typically require substantially larger model parameters or architectural modifications.

\section{Related Work}

The evolution of reasoning capabilities in large language models has progressed from early chain-of-thought (CoT) prompting techniques \citep{wei2022chain,zhang2023multimodal} to process reward models \citep{lightman2023let,uesato2022solving}. Advanced reasoning models like OpenAI's o1 and DeepSeek-R1 employ internal test-time scaling, generating extended chain-of-thought sequences before producing final answers \citep{guo2025deepseek,jaech2024openai}. However, Large Reasoning Models exhibit the "overthinking phenomenon" where longer reasoning sequences improve performance but introduce significant computational overhead \citep{sui2025stop,chen2024not}. To address efficiency concerns, recent work has explored adaptive reasoning frameworks that dynamically adjust computational allocation based on task complexity. Several approaches focus on mode switching mechanisms, including Self-Route for automatic capability-based routing, ThinkSwitcher for systematic fast-slow thinking transitions, and Qwen3's seamless switching between thinking and non-thinking modes \citep{he2025self,liang2025thinkswitcher,yang2025qwen3}. Other methods target reasoning length optimization, such as AutoL2S which combines long and short chain-of-thought data adaptively, and TLDR for controlled reasoning compression \citep{luo2025autol2s,li2025tl}. Additional innovations include AdaptThink's learning-based reasoning engagement and various hybrid approaches like Llama-Nemotron, KAT-V1, and AdaCoT that integrate multiple adaptive strategies \citep{zhang2025adaptthink,bercovich2025llama,zhan2025kat,lou2025adacot}. Despite these advances in adaptive reasoning strategies, existing methods primarily operate at the problem level without fine-grained resource allocation within sub-problems. SCALE addresses this gap through fine-grained resource allocation at the sub-problem level, achieving superior computational efficiency through selective cognitive resource distribution.

\section{Conclusion}

This work addresses a fundamental bottleneck in test-time compute scaling for mathematical reasoning: the inefficient uniform resource allocation that prevents effective scaling despite increasing computational budgets. Our proposed SCALE framework overcomes these limitations through selective resource allocation inspired by dual-process theory. By decomposing problems into sequential reasoning sub-problems, assessing their difficulty, and dynamically assigning appropriate processing modes, SCALE concentrates computational resources where they can maximize impact. Extensive experiments demonstrate SCALE's effectiveness, achieving substantial accuracy improvements while demonstrating superior resource utilization efficiency compared to uniform scaling approaches.

\section{Acknowledgments}
We would like to thank all reviewers for their insightful comments and suggestions to help improve the paper. This work was supported by the Research Grants Council of Hong Kong (GRF No. 15209724). 

% This work was also partially funded by the National Natural Science Foundation of China (62476168) and SII.

\bibliography{aaai2026}

\end{document}